%% file: bare_jrnl_new_sample4.tex
\documentclass[lettersize,journal]{IEEEtran}
\usepackage{amsmath,amsfonts}
\usepackage{algorithmic}
\usepackage{algorithm}
\usepackage{array}
\usepackage[caption=false,font=normalsize,labelfont=sf,textfont=sf]{subfig}
\usepackage{textcomp}
\usepackage{stfloats}
\usepackage{url}
\usepackage{verbatim}
\usepackage{graphicx}
\usepackage{cite}
\usepackage{booktabs}
\usepackage[utf8]{inputenc} 
\usepackage[T1]{fontenc}    
\usepackage{hyperref}       
\usepackage{nicefrac}       
\usepackage{microtype}      
\usepackage{booktabs}
\usepackage{tabularx}
\usepackage{array}
\usepackage{multirow}

\usepackage[table]{xcolor}
\usepackage{hyperref}
\usepackage{orcidlink}
\usepackage[table]{xcolor}

\definecolor{FISRlight}{HTML}{F1F6FC}
\definecolor{FISRppLight}{HTML}{E3EDF8}

\definecolor{FISRteal}{HTML}{F0F7F5}
\definecolor{FISRppTeal}{HTML}{E0EFEB}

\definecolor{FISRgreen}{HTML}{F2F7EF}
\definecolor{FISRppGreen}{HTML}{E5F0DF}

\definecolor{FISRpurple}{HTML}{F5F1F9}
\definecolor{FISRppPurple}{HTML}{EBE4F3}

\begin{document}

\title{\texttt{STAR++}: A Physical-Aware Benchmark and Framework for Astronomical Super-Resolution across Multiple Filters}

\author{Guohang~Zhuang\textsuperscript{*}\textsuperscript{\orcidlink{0000-0002-4744-7841}}, 
        Jiaze Li\textsuperscript{*}\textsuperscript{\orcidlink{0000-0002-3351-015X} }, 
        Kuo-Cheng~Wu\textsuperscript{\orcidlink{0009-0008-0474-8354}},
        Hao~Du\textsuperscript{\orcidlink{0009-0006-6685-7914}}, 
        Jinyang~Huang\textsuperscript{\orcidlink{0000-0001-5483-2812}}, 
        Xiang~Zhang\textsuperscript{\orcidlink{0000-0003-0413-6135}}, \\
        Wanli~Ouyang\textsuperscript{\orcidlink{0000-0002-9163-2761}},~\IEEEmembership{Senior Member,~IEEE}, 
        and Yan~Lu\textsuperscript{\orcidlink{0009-0002-1449-5174}}

\thanks{Guohang Zhuang and Jinyang Huang are with the School of Computer Science and Information Engineering, Hefei University of Technology, Hefei 230026, China (e-mail: guohang\_zhuang@mail.hfut.edu.cn; hjy@hfut.edu.cn). Guohang Zhuang is also with Shanghai Artificial Intelligence Laboratory, Shanghai 200240, China (e-mail: zhuangguohang@pjlab.org.cn).}
\thanks{Jiaze Li is with the University of Science and Technology of China, Hefei (jz\_li@mail.ustc.edu.cn).}
\thanks{Hao Du, Kuo-Cheng Wu, Wanli Ouyang, and Yan Lu are with The Chinese University of Hong Kong, Hong Kong (e-mail: dh97@mail.ustc.edu.cn; 12guocheng@gmail.com; wlouyang@ie.cuhk.edu.hk; luyan@pjlab.org.cn).}
\thanks{Wanli Ouyang and Yan Lu are also with Shanghai Artificial Intelligence Laboratory, Shanghai 200240, China.}
\thanks{Xiang Zhang is with the School of Cyber Security, Tianjin University, Tianjin, China (e-mail: zhangxiang@ieee.org).}
\thanks{This work was done during Guohang Zhuang's internship at Shanghai Artificial Intelligence Laboratory.}
\thanks{* Guohang Zhuang and Jiaze Li contributed equally to this work.}
\thanks{This work was supported by the JC STEM Lab of AI for Science and Engineering, funded by The Hong Kong Jockey Club Charities Trust, the Research Grants Council of Hong Kong (Project No. CUHK14213224).}

}


\markboth{Journal of \LaTeX\ Class Files,~Vol.~14, No.~8, August~2021}%
{Shell \MakeLowercase{\textit{et al.}}: A Sample Article Using IEEEtran.cls for IEEE Journals}



\maketitle

\input{sec/0_abstract}
\begin{IEEEkeywords}
Super-Resolution, Benchmark, Flux Consistency, Cross-Filter Generalization.
\end{IEEEkeywords}
\input{sec/1_introduction}

\input{sec/2_related_works}
\input{sec/3_method}
\input{sec/4_Flux-guided_model_optim}
\input{sec/5_experiment}

\input{sec/6_conclusion}

\input{sec/7_limitaions_ack}
\bibliographystyle{unsrt}
\bibliography{ref} 

\clearpage
\newpage

\end{document}

%% file: sec/0_abstract.tex
\begin{abstract}


Astronomical super-resolution (ASR) aims to reconstruct low-resolution cosmic observations from the telescope using multi-filter, enabling accurate reconstruction of star fields for astrophysical analysis. However, developing a physics-aware ASR framework faces challenges from two perspectives:
1) Data view: no comprehensive or large-scale benchmark supports the development of ASR methods for field-level reconstruction, especially on unseen filter settings.
2) Method view: Existing ASR methods often fail to maintain flux consistency for stellar sources across different spatial resolutions. Moreover, ASR requires the reconstruction of high-frequency components, such as accurate star locations and profiles, leading to higher spatial ill-posedness than natural-image SR.
To tackle these problems, we first introduce STAR++, a large-scale benchmark comprising approximately 1.7 million flux-consistent image pairs across 36 filters, derived from Hubble Space Telescope (HST) observations.
STAR++ is constructed through a flux-preserving data generation pipeline that ensures photometric consistency across different spatial resolutions, and it provides a novel Flux Error (FE) metric to evaluate ASR models from a physical perspective. 
To address these method-level challenges, we propose a Flux-Invariant Super-Resolution Plus~(FISR++) model that generalizes to unseen filters while accurately reconstructing flux-consistent high-resolution images. Except for the flux-preserving modules, FISR++ further introduces two novel components. Specifically, a Filter-Aware Adapter~(FAA) incorporates filter metadata to adapt image representations to varying spectral and optical conditions. In addition, a Spatial Ambiguity Disentangle Loss~(SADL) explicitly models spatial ambiguity as a pixel shift uncertainty, mitigating the smoothing of learned conditional mean reconstructions caused by this ambiguity.
Extensive experiments demonstrate that FISR++ consistently outperforms existing ASR methods in reconstruction quality and flux accuracy, while exhibiting stronger generalization to unseen filters, highlighting its effectiveness for photometrically faithful ASR.

\end{abstract}

%% file: sec/1_introduction.tex
\section{Introduction}
\label{sec:intro}

Higher resolution is critical to astronomical observation, as it reveals finer astrophysical structures and supports more precise scientific measurements~\cite{labeyrie1977ii,bryant2021episodic,faaique2024overview}. Achieving higher spatial resolution traditionally relies on advances in telescopes and observing systems, often at substantial instrumental and operational cost~\cite{mccray2006giant,mccray2000large}. Super-resolution~(SR) offers a complementary computational approach to reconstruct high-resolution images from lower-resolution observations~\cite{dong2015image,huang2015single,lim2017enhanced}, and has shown promise in astronomical imaging~\cite{marsh2004enhancement,puschmann2005super}. However, astronomical images contain physically meaningful information, such as the flux of celestial sources. Therefore, developing a physics-aware ASR framework is essential for accurately reconstructing star fields while preserving their physical fidelity.

Physics-aware ASR requires cross-filter generalization, flux consistency~(i.e., consistent stellar brightness across spatial resolutions), and accurate reconstruction of stellar structures. However, existing ASR datasets and methods still face limitations from two perspectives to achieve the aforementioned goals.
1). \textbf{Data perspective}: Existing ASR benchmarks cover a limited number of filters and lack systematic evaluation on unseen filters~\cite{miao2024astrosr,shan2025galaxy,luo2025cross}, as shown in Tab.~\ref{tab:dataset-comparison}. Moreover, existing ASR networks primarily rely on image inputs without explicitly incorporating filter-specific physical properties such as wavelength and bandwidth, limiting their adaptability to different observation conditions.
2). \textbf{Method perspective}: It mainly includes two challenges of flux inconsistency and spatial ill-posedness. (i)~\textbf{Flux inconsistency}: In the real world, telescopes under different observation resolutions have a flux consistency relation~\cite{gordon2022james,smith2009absolute}. Specifically, although a celestial object has different levels of distortion under low resolutions, it almost has the same total flux as in high resolutions because of the telescope imaging principle~\cite{gordon2022james,bohlin2019hubble}. However, existing datasets have significant drifts from this property because they directly use simple interpolation~\cite{song2024improving,reddy2024difflense}, suitable for natural images but conflicts with astronomical observations. (ii)~\textbf{Spatial ill-posedness}: Unlike natural image super-resolution, which can use surrounding context to restore large objects, ASR must recover isolated, tiny point sources like stars. Without contextual clues, a single low-resolution pixel can correspond to many possible high-resolution stellar positions and profiles, making the recovery severely ill-posed. To minimize error under this spatial ambiguity, common likelihood training simply averages all candidate positions, which flattens the peak intensity, leading to over-smoothed restoration and inaccurate source flux.

\begin{figure*}
  \centering
  \label{fig:intro}\includegraphics[width=1.5\columnwidth]{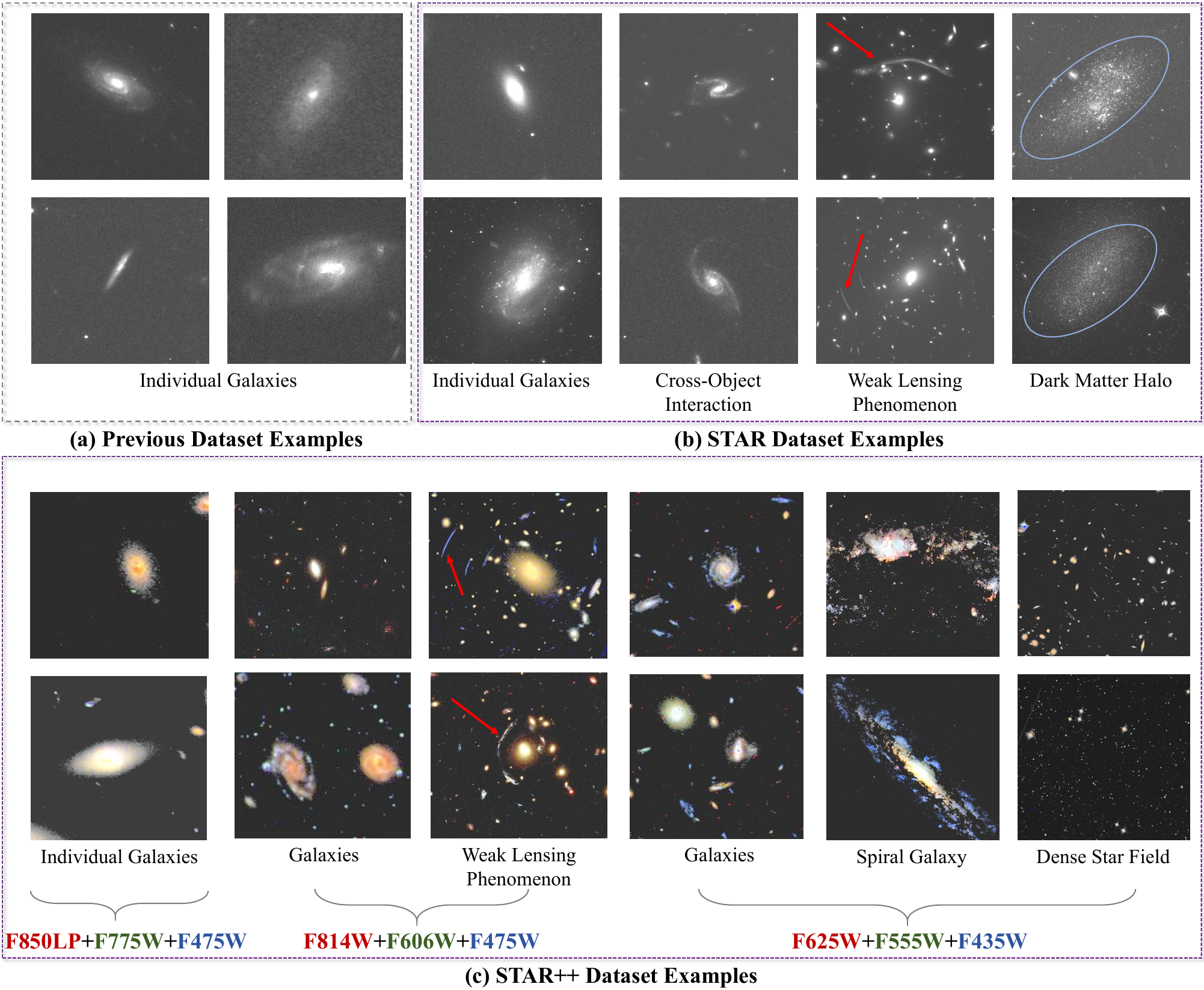}
  \caption{Comparison of previous datasets and ours, highlighting the richer astronomical structures and phenomena captured by STAR++, such as cross-object interactions, weak lensing, Dense star fields, and spiral galaxies. The STAR++ examples are visualized as RGB composites, with \textcolor{red}{red}, \textcolor{green}{green}, and \textcolor{blue}{blue} filter labels indicating the filters mapped to the corresponding color channels.}
\end{figure*}

To support comprehensive, large-scale, multi-filter ASR research across diverse astronomical fields, we introduce \textbf{STAR++}, a flux-consistent benchmark for field-level astronomical super-resolution.
It extends STAR~\cite{star} from one filter to 36 filters spanning the 222.4--906.0~nm band. STAR++ contains nearly 1.7 million flux-consistent LR-HR pairs, generated from 2,019 Hubble Space Telescope~(HST)~\cite{bohlin2019hubble} science images. The source images cover diverse astronomical fields, covering a large range of star fields and on average containing 55 celestial objects per image and complex scenarios including multiple celestial objects, cross-object interactions, and weak lensing phenomena, as illustrated in Fig.~\ref{fig:intro}. STAR++ retains the flux-consistent data generation pipeline of STAR to preserve the photometric properties of astronomical observations across resolutions. Further, we provide Flux Error (FE) to evaluate reconstruction quality from a physical perspective.

Building on STAR++, we propose a novel field-level ASR model, Flux-Invariant Super-Resolution Plus~(FISR++), to improve cross-filter generalization and reconstruction accuracy while maintaining flux consistency. In addition to the flux-preserving Flux Guidance Generation~(FGG) and Flux Guidance Controller~(FGC) modules and the Flux Consistency Loss~(FCL)\cite{star}, FISR++ incorporates two novel modules, namely a Filter-Aware Adapter~(FAA) and Spatial Ambiguity Disentangle Loss~(SADL). Specifically, the FAA incorporates filter metadata, allowing the model to recognize filter and adapt filter-specific knowledge to restore high resolution image across different filters. This filter-aware design enables the model to adapt its super-resolution patterns according to the filter condition, thereby improving its generalization to unseen filters. In addition, we propose a Spatial Ambiguity Disentanglement Loss~(SADL) to accurately restore stellar positions and profiles. During training, SADL explicitly models spatial ambiguity as pixel-shift uncertainty and separates it from image reconstruction. Learning this uncertainty independently keeps the model from averaging across candidate positions, so it can pinpoint exact stellar peaks and preserve sharp profiles. At inference, we simply drop the uncertainty term to produce clean, sharp high-resolution images.

In summary, the key contributions of this paper are as follows:

(1) \textbf{STAR++ Benchmark}: We present STAR++, a multi-filter, flux-consistent ASR benchmark comprising approximately 1.7 million LR–HR pairs with 36 HST filters. 

(2) \textbf{Flux-Invariant Super Resolution Plus~(FISR++) Model}: We propose FISR++, extending FISR with a Filter-Aware Adapter~(FAA) that incorporates physical filter priors and a Spatial Ambiguity Disentangle Loss that accounts for pixel-wise uncertainty and sub-pixel misalignment.

(3) Extensive experiments demonstrate that FISR++ achieves state-of-the-art performance on STAR++, with a PSNR of 35.42 dB and a Flux Error (FE) of 0.4341 on unseen filters. Compared with the conference-version FISR, FISR++ improves PSNR by 1.1 dB and reduces FE by 0.0963.

\textit{Difference from the conference paper:} The preliminary version of this manuscript is conference paper STAR~\cite{star}. Compared with the conference version, the improvements of the STAR++ and FISR++ are listed below: 1) We expand STAR from a single filter to 36 HST filters, increasing the dataset from 54,738 to approximately 1.7 million flux-consistent LR–HR pairs.  2) The proposed Filter-Aware Adapter~(FAA) incorporate filter metadata into image representations. We also adopt an asinh intensity representation to compress the dynamic range and better balance the contributions of bright and faint sources during training. 3) We introduce a Spatial Ambiguity Disentangle Loss that explicitly models pixel shifts and pixel-wise reconstruction uncertainty to improve the positional accuracy of reconstructed high-frequency structures and reduce spatial ambiguity. 4) We conduct additional experiments to evaluate generalization to unseen filters and compare the performance of models trained on STAR and STAR++. Furthermore, FISR++ achieves a PSNR of 35.42 dB and a Flux Error (FE) of 0.4341 on STAR++, representing a 1.1 dB improvement in PSNR and a 0.0963 reduction in FE over FISR, thereby demonstrating state-of-the-art performance.

\begin{table*}[t]
\centering
\caption{Comparison of existing ASR datasets.}
\label{tab:dataset-comparison}
\resizebox{\textwidth}{!}{%
\begin{tabular}{c c c c c c c c}
\toprule
\textbf{Dataset} &
\textbf{Size} &
\textbf{Type} &
\textbf{\# Filters} &
\textbf{Downsampling} &
\textbf{Field-level} &
\textbf{Flux Consistency} &
\textbf{Unseen-filter Eval.} \\
\midrule

AstroSR~\cite{miao2024astrosr}
& 2,000 & Galaxy & 5 & $\times$ & $\times$ & $\times$ & $\times$ \\

Shan et al.~\cite{shan2025galaxy}
& 9,383 & Galaxy & 1 & $\times$ & $\times$ & $\times$ & $\times$ \\

Song et al.~\cite{song2024improving}
& 1,597 & Solar & 1 & $\checkmark$ & $\times$ & $\times$ & $\times$ \\

DiffLense~\cite{reddy2024difflense}
& 2,880 & Galaxy & 1 & $\times$ & $\times$ & $\times$ & $\times$ \\

Luo et al.~\cite{luo2025cross}
& 17,000 & Galaxy & 3 & $\times$ & $\times$ & $\times$ & $\times$ \\

Li et al.~\cite{li2022self}
& 14,604 & Galaxy & 3 & $\checkmark$ & $\times$ & $\times$ & $\times$  \\

\rowcolor[gray]{0.9}
STAR
& 54,738 & Star field & 1
& $\checkmark$ & $\checkmark$ & $\checkmark$ & $\times$ \\

\rowcolor[gray]{0.9}
\textbf{STAR++}
& \textbf{1,697,816} & Star field & \textbf{36}
& $\checkmark$ & $\checkmark$ & $\checkmark$ & $\checkmark$ \\
\bottomrule
\end{tabular}%
}
\end{table*}

%% file: sec/2_related_works.tex
\section{Related Works}
\label{sec:rela_works}

\subsection{Astronomical Super-Resolution Benchmarks}
Astronomical super-resolution~(ASR) aims to reconstruct high-resolution astronomical observations from low-resolution images. Developing reliable ASR models requires large-scale training data, which has motivated the recent development of several ASR datasets and benchmarks. AstroSR~\cite{miao2024astrosr} constructed 2,000 paired galaxy images from overlapping SDSS and HSC observations for cross-survey galaxy super-resolution. Shan et al.~\cite{shan2025galaxy} further collected 9,383 galaxy image pairs for diffusion-based reconstruction, while Song et al.~\cite{song2024improving} and DiffLense~\cite{reddy2024difflense} extended ASR to solar images and gravitational-lensing observations, respectively. Luo et al.~\cite{luo2025cross} and Li et al.~\cite{li2022self} explored cross-survey ASR with datasets containing 17,000 and 14,604 image pairs, respectively. Compared with these datasets, STAR~\cite{star} introduced 54,738 flux-consistent HST star-field pairs for field-level ASR, covering multiple celestial objects and complex phenomena such as cross-object interactions and weak lensing while preserving physical fidelity. More recently, StelLens~\cite{ouyang2025imaging} constructed a large-scale paired SDSS-HST dataset for learning ground-to-space astronomical reconstruction. FluxFlow~\cite{FluxFlow} introduced a DESI-HST benchmark with 19,500 image pairs and realistic atmospheric degradation. However, despite these advances in dataset scale and astronomical targets, existing ASR benchmarks still cover only a limited number of predefined spectral filters, leaving cross-filter and unseen-filter generalization insufficiently studied.

\subsection{Super-Resolution Techniques}
Super-resolution (SR) techniques aim to reconstruct high-resolution (HR) images from low-resolution (LR) observations. Traditional approaches, including interpolation, deconvolution, and sparse representation~\cite{zeyde2010single}, recover image details through predefined degradation models or learned statistical relationships. Deep learning has substantially advanced SR through convolutional neural networks (CNNs)~\cite{dong2015image,lim2017enhanced,zhang2018image}, while generative adversarial networks (GANs) have improved perceptual reconstruction quality~\cite{ledig2017photo,wang2021real,chan2022glean}. Transformer-based methods further exploit attention mechanisms to capture long-range dependencies and recover fine image structures~\cite{liang2021swinir,chen2023activating,zhang2022efficient,lu2022transformer}. More recently, diffusion-based methods have employed iterative denoising processes to generate high-quality reconstructions~\cite{saharia2022image,wang2024exploiting,rombach2022high}. Although these techniques have achieved substantial improvements in image reconstruction, visual similarity and perceptual quality alone do not guarantee the physical fidelity required for astronomical analysis.

\subsection{Astronomical Super-Resolution Methods}
Astronomical image reconstruction has traditionally relied on deconvolution~\cite{puschmann2005super,starck2002deconvolution}, which models the point spread function (PSF)~\cite{breckinridge2015polarization}, and multi-exposure reconstruction~\cite{hirsch2011online,zibetti2007optical}, which combines information from multiple observations. Deep learning has subsequently enabled reconstruction of various astronomical targets, including galaxies~\cite{schawinski2017generative}, solar structures~\cite{baso2018enhancing}, and X-ray sources~\cite{sweere2022deep}. Beyond individual celestial objects, STAR~\cite{star} introduced field-level ASR for complex star fields containing multiple objects with varying brightness. Recent astronomy-specific methods have further incorporated physical information into image reconstruction. STAR~\cite{star} introduced flux-guided reconstruction, while StelLens~\cite{ouyang2025imaging} explored observation-physics-driven ground-to-space reconstruction. FluxFlow~\cite{FluxFlow} employed flow matching with observation uncertainty to improve astronomical reconstruction. Nevertheless, robust reconstruction across diverse and previously unseen filters remains challenging, particularly when observation characteristics vary across filters. Moreover, small positional deviations between reconstructed and reference images can introduce substantial pixel-wise errors, limiting the accurate recovery of stellar structures.

\subsection{Flux Consistency in Astronomical Image Processing}
Flux consistency, ensuring that the total light intensity (flux, or photons received per unit area) of celestial objects in processed images matches original observations, is a cornerstone of astronomical image analysis, underpinning reliable photometry and stellar population studies~\cite{van1967ubv,bessell2005standard}. Historical efforts prioritized flux consistency to preserve measurement accuracy in star clusters and galaxies. The ground-based survey telescope SDSS also follows this principle~\cite{padmanabhan2008improved}. However, the complexity of star fields, with diverse brightness and overlapping objects, poses ongoing challenges. Our work advances this field with STAR++, a large-scale star field dataset ensuring flux-consistent image pairs, and novel flux consistency constraints, enhancing the scientific reliability of star field analyses.

\begin{table*}[t]
\centering
\footnotesize
\caption{
Dataset statistics, data splits, and filter characteristics of STAR++, including pivot wavelength and fractional bandwidth. Unseen filters are divided into interpolation and extrapolation subsets according to their pivot wavelengths relative to the training wavelength range.}
\label{tab:starpp_overview_split}

\setlength{\tabcolsep}{5pt}
\renewcommand{\arraystretch}{1.12}

\begin{tabular*}{\textwidth}{
    @{\extracolsep{\fill}}llccrr@{}
}
\toprule

\multicolumn{6}{@{}l@{}}{
    \textbf{Imagers:}
    WFC3/UVIS (1,106 images);
    ACS/WFC (913 images)
} \\


\multicolumn{6}{@{}l@{}}{
    \textbf{Sky coordinates:}
    RA $0.07^{\circ}$\textasciitilde$359.94^{\circ}$;
    Dec $-89.13^{\circ}$\textasciitilde$87.41^{\circ}$
} \\

\midrule

\textbf{Split}
& \textbf{Filter(s)}
& \textbf{Pivot. $\lambda$ (nm)}
& \textbf{Frac. BW}
& \textbf{Images}
& \textbf{LR--HR pairs} \\

\midrule

Training (seen)
& 29 filters & 270.3--843.7 & 0.004--0.352
& 1,008 & 922,586 \\

Validation (seen)
& 29 filters & 270.3--843.7 & 0.004--0.352
& 237 & 162,593 \\

\midrule

\multirow{5}{*}{Test (unseen-interp.)}
& F435W & 432.9 & 0.069 & 82 & 76,019 \\
& F555W & 533.6 & 0.082 & 93 & 47,169 \\
& F606W & 591.2 & 0.113 & 379 & 351,326 \\
& F625W & 629.1 & 0.068 & 52 & 48,623 \\

\cmidrule(lr){2-6}

& \textit{Subtotal}
& 432.9--629.1 & 0.068--0.113
& 606 & 523,137 \\

\midrule

\multirow{4}{*}{Test (unseen-extrap.)}
& F218W & 222.4 & 0.056 & 55 & 3,500 \\
& F225W & 236.5 & 0.074 & 53 & 38,536 \\
& F850LP & 906.0 & 0.057 & 60 & 47,464 \\

\cmidrule(lr){2-6}

& \textit{Subtotal}
& 222.4--906.0 & 0.056--0.074
& 168 & 89,500 \\

\midrule

\textbf{Total}
& \textbf{36 filters}
& 222.4--906.0
& 0.004--0.352
& \textbf{2,019}
& \textbf{1,697,816} \\

\bottomrule
\end{tabular*}
\end{table*}


%% file: sec/3_method.tex
\section{STAR++}
\label{sec:method}

Following natural SR works, we construct cross-resolution image pairs by downsampling high-resolution images. We choose Hubble Space Telescope~(HST)\footnote{\url{https://www.stsci.edu/hst}} survey data as our high-resolution images due to its widely recognized data quality and rich historical data accumulation.
Given a high-resolution HST image, we first apply a point spread function~(PSF)~\cite{breckinridge2015polarization} kernel to simulate the optical blurring effects caused by low-quality telescopes and atmospheric turbulence. 

Next, we downsample the image by a factor of $s$ using a flux-conserving scheme. Finally, since both HR and LR images have large spatial dimensions, we divide them into smaller sub-images to facilitate model training. This pipeline generates physically consistent cross-resolution pairs of images, crucial for robust super-resolution model training. The statistics of the resulting dataset are summarized in Tab.~\ref{tab:starpp_overview_split}.

\subsection{High-resolution Data Collection}

The HST is a space-based observatory designed to capture high-resolution astronomical images across a wide range of wavelengths, from ultraviolet to near-infrared. It provides two kinds of data, including \texttt{calibration} and \texttt{science}. Calibration data is used to correct instrumental effects while the science has a verified quality for scientific research. 
So we choose scientific data due to its high and reliable quality.

The \texttt{science} data consist of images acquired by various imaging instruments onboard HST. We select two representative high-resolution imagers, including the Advanced Camera for Surveys Wide Field Channel (ACS/WFC)~\cite{ford2003overview} and the Wide Field Camera 3 Ultraviolet-Visible Channel (WFC3/UVIS)~\cite{dressel2010wide}. ACS/WFC provides a wide field of view ($202''\times202''$) and high sensitivity over optical wavelengths (370-1100 nm), while WFC3/UVIS provides a field of view ($162''\times162''$) with broader ultraviolet-to-red coverage (200-1000 nm). Their high spatial resolution, large fields of view, and complementary spectral coverage make them well suited for constructing a large-scale multi-filter ASR benchmark.

In practice, HST observations are acquired through multiple filters with distinct wavelength-dependent transmission profiles. Since celestial sources exhibit wavelength-dependent emission, their observed intensities vary across filters, making multi-filter coverage essential for evaluating cross-filter generalization in ASR.  STAR++ incorporates 36 HST filters to substantially increase spectral diversity. These filters have pivot wavelengths ranging from 222.4 to 906.0 nm and include wide, medium, narrow, long-pass, and extremely wide filters. As shown in Tab.~\ref{tab:starpp_overview_split}, we reserve seven filters for testing to evaluate generalization, and their corresponding central wavelengths are not represented among the filters used for training. Such diverse filter configurations enable STAR++ to cover multiple filter observation conditions and support systematic analysis of cross-filter generalization.

\begin{figure*}
  \centering
  \label{figflux}
  \includegraphics[width=1.5\columnwidth]{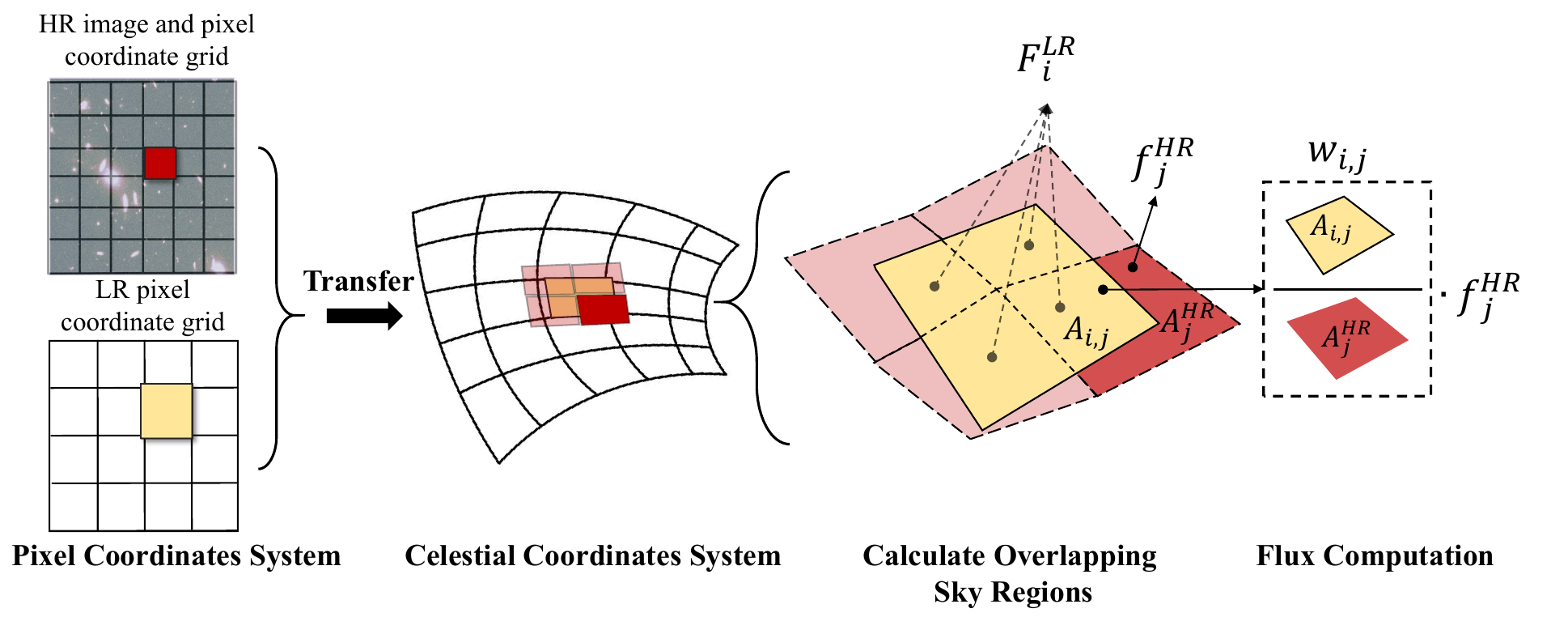}
   \caption{Schematic diagram of the flux-consistent downsampling process. The workflow illustrates the transformation of HR and LR image pixels into the celestial coordinate system, the computation of overlapping sky regions between HR and LR receptive fields, and the flux calculation for LR pixels using weighted contributions from overlapping HR pixels.}
    \label{figflux}
\end{figure*}

\subsection{PSF Blurring}
Different resolution devices share different PSF blurring. To simulate this phenomenon and increase training data diversity, we adopt two representative PSF models. The first is the Gaussian PSF~\cite{koenderink1987representation}, a simple model widely used to approximate blur caused by atmospheric turbulence or instrumental imperfections in astronomical observations~\cite{hardy1998adaptive, roddier1981v}:
\begin{equation}
\text{PSF}_{\text{Gaussian}}(x, y) = \exp\left(-\frac{x^2 + y^2}{2\sigma^2}\right).
\label{eq:gaussian_psf}
\end{equation}
The second is the Airy PSF, a device-specific kernel widely used in astrophysics. It captures diffraction effects from a telescope's circular aperture and describes the instrumental effect that occurs when a telescope resolution changes, making it suitable for space-based instruments like HST~\cite{scoville2007cosmos}:
\begin{equation}
\text{PSF}_{\text{Airy}}(r) = \left[ \frac{2 J_1(kr)}{kr} \right]^2.
\label{eq:airy_psf}
\end{equation}

In these two models, $\sigma$ and $r$ serve as adjustable parameters to control the spread of the blur by modulating the energy dispersion of the filter. In the Gaussian PSF, a larger $\sigma$ leads to a less concentrated signal with greater energy spread across the filter, while in the Airy PSF, $r$ governs the radial extent of energy distribution due to diffraction. We define these parameters based on the telescope's observed blur characteristics, following Schawinski et al.~\cite{schawinski2017generative}, who used the observed blur to set the PSF parameters for a realistic simulation of hardware-specific degradation effects. 

With these two PSFs, we define a blurring setting as a random selection between Airy and Gaussian, each chosen with equal probability, to simulate more complex scenarios. After operating PSF blurring, we perform a flux consistency downsampling scheme to obtain realistic low-resolution counterparts, which we detail in the next section.




\subsection{Flux Consistency Downsampling}
The flux consistency relation in real-world telescopic observations stems from the telescope imaging principle. The value of each pixel in observational data corresponds to the photon flux captured by its corresponding CCD pixel. Consequently, when imaging the same celestial region, a single CCD pixel in lower-resolution instruments covers a larger spatial area—equivalent to integrating photons from multiple high-resolution CCD pixels. Based on this mechanism, Flux-Consistent Downsampling is implemented in three steps: computing the image-plane coordinates, transferring them to celestial coordinates, and calculating the flux of each LR pixel according to the overlap between HR and LR pixel regions.

\textbf{Computing image plane coordinates: }For a given high-resolution (HR) image with resolution $H\times W$ and downsampling rate $s$, we first generate the downsampled low-resolution (LR) grid after valid padding or cropping so that paired HR and LR regions keep an integer scale relation. The resulting LR image size is $\frac{H}{s}\times \frac{W}{s}$. These HR and LR pixel coordinates define the image-plane grids used for WCS reprojection.

\textbf{Transfer pixels to sky: }For HR and LR pixel coordinates, we transfer them into the celestial coordinate system as: $(u,v)\rightarrow(ra,dec)$, where $(u, v)$ is a coordinate in the image plane while $(ra,dec)$ is the longitude and latitude coordinates of the Earth. Note that, each pixel is not an ideal point and actually a rectangle on the image plane. After the mapping, it becomes a quadrilateral surface of the celestial coordinate system. The physical meaning of this quadrilateral surface is the sky area covered by a pixel, denoted as the receptive field here. For the $i$-th pixel of the LR image and the $j$-th pixel of the HR image, we calculate and denote the area value of their receptive field as $A^{LR}_i$ and $A^{HR}_j$.

The transformation process in the aforementioned process is implemented by the telescope calibration information carried by the high-resolution (HR) image, which could be interpreted as camera intrinsic and extrinsic parameters of the telescope. 

\textbf{Low-resolution image Flux Computation: } The previous steps essentially transferred HR and LR image plane grids into two surface meshes in the celestial coordinate system, as shown in Fig.~\ref{figflux}. Obviously, the average receptive field of the LR image is larger than the HR one because the LR pixel corresponds to larger regions, leading to an LR pixel covers multiple HR pixels in the sky. To compute the flux of the $i$-th LR pixel, we first identify the set of HR pixels $S^o_i$ whose receptive fields overlap with that of the $i$-th LR pixel, i.e., $S^o_i = \{ j \mid A^{HR}_j \cap A^{LR}_i \neq \emptyset \}$. This set represents all HR pixels whose sky areas contribute to the $i$-th LR pixel's flux. The flux of the $i$-th LR pixel, $F^{LR}_i$, is then computed by summing the weighted contributions from all overlapping HR pixels:

\begin{equation}
\label{eq:flux_lr}
F^{LR}_i = \sum_{j \in S^o_i} w_{i,j} \cdot f^{HR}_j,
\end{equation}
where $f^{HR}_j$ is the flux of the $j$-th HR pixel, and $w_{i,j}$ is the weight representing the fractional contribution of the $j$-th HR pixel to the $i$-th LR pixel.

The weight $w_{i,j}$ is calculated as:
\begin{equation}
w_{i,j} = \frac{A_{i,j}}{A^{HR}_j},
\end{equation}
where $A_{i,j}$ is the overlapping sky area between the $i$-th LR pixel and the $j$-th HR pixel, representing their shared quadrilateral patch in the celestial coordinate system, and $A^{HR}_j$ is the total sky area covered by the $j$-th HR pixel's receptive field. To better understand the role of this weight in flux computation, we substitute $w_{i,j}$ into Equation~\eqref{eq:flux_lr}, transforming the contribution term as follows. The flux contribution from the $j$-th HR pixel to the $i$-th LR pixel is $w_{i,j} \cdot f^{HR}_j$, where $f^{HR}_j$ is the flux of the $j$-th HR pixel. Substituting $w_{i,j} = \frac{A_{i,j}}{A^{HR}_j}$ into this term, we obtain:
\begin{equation}
w_{i,j} \cdot f^{HR}_j = \left( \frac{A_{i,j}}{A^{HR}_j} \right) \cdot f^{HR}_j = A_{i,j} \cdot \frac{f^{HR}_j}{A^{HR}_j}.
\end{equation}
Here, $\frac{f^{HR}_j}{A^{HR}_j}$ represents the flux density of the $j$-th HR pixel, i.e., the photon count per unit sky area, as recorded by the telescope's CCD sensor over the receptive field area $A^{HR}_j$. Thus, $A_{i,j} \cdot \frac{f^{HR}_j}{A^{HR}_j}$ is the flux contributed by the $j$-th HR pixel over the overlapping area $A_{i,j}$, ensuring that the contribution is proportional to the shared sky area between the LR and HR pixels. This approach preserves the total photon flux during downsampling, maintaining flux consistency across resolutions.

\subsection{HR Image Subdivision and Data Splitting}
The previous flux-consistent downsampling scheme provides flux-consistent HR-LR image pairs. To further optimize model training effectiveness, we divide the HR and LR wide field images into smaller sub-images. Because astronomical images contain some outlier regions~(have NaN values) due to geometric calibration in DrizzlePac\cite{gonzaga2012drizzlepac} processing, we retain only patches with >80\% valid regions containing stellar features. To prevent data leakage, we split the dataset by sky region, ensuring that patches from the same sky region are assigned to the same split. We further verify that the training, validation, and test sets have no overlapping sky coverage or shared raw exposures. Through these processing steps, we obtain a large set of high-quality HR-LR image pairs and construct the STAR++ dataset for training and evaluating astronomical super-resolution models.

\subsection{Flux Error}
STAR++ provides a Flux Error (FE) to evaluate flux consistency between a ground truth and its corresponding prediction.
The FE measures the flux value gap for each object, so its computation process is based on astrophysical photometry. The basic process of the photometry is deriving flux by detection. We follow this idea. For a given ground-truth and predicted image pair, we compute the FE as following two steps: 
\textbf{1)}. For the ground-truth image, we use the Starfinder toolkit~\cite{barbary2018sep} to detect celestial objects and obtain their parameters. Then, we derive the flux of each object by a widely-used elliptical photometry method~\cite{barbary2018sep}.
\textbf{2)}. For predicted images, we do not operate object detection but directly use detection results from the ground-truth image because they provide reliable object catalogs covering both strong and weak sources. The following photometry is the same as the ground truth. After these two steps, we have a two set of flux values, denoted as $\{v^1_{pred},v^2_{pred}...,v^N_{pred}\}$ and $\{v^1_{gt},v^2_{gt}...,v^N_{gt}\}$ where $N$ is the number of detected objects. The FE is computed by the following: 
\begin{equation}
\text{FE} = \frac{1}{N} \sum_{i=1}^N \left| v^i_{gt} - v^i_{pred} \right|.
\end{equation}


Lower FE means higher flux consistency. Since the flux value is related to the object shape and the pixel flux in object regions, this metric could reflect the geometric shape consistency for each object in the reconstruction image, physically informed flux consistency and weak source object reconstruction quality simultaneously.  

\begin{figure*}
  \centering
  \label{fig:method}
  \includegraphics[width=2.0\columnwidth]{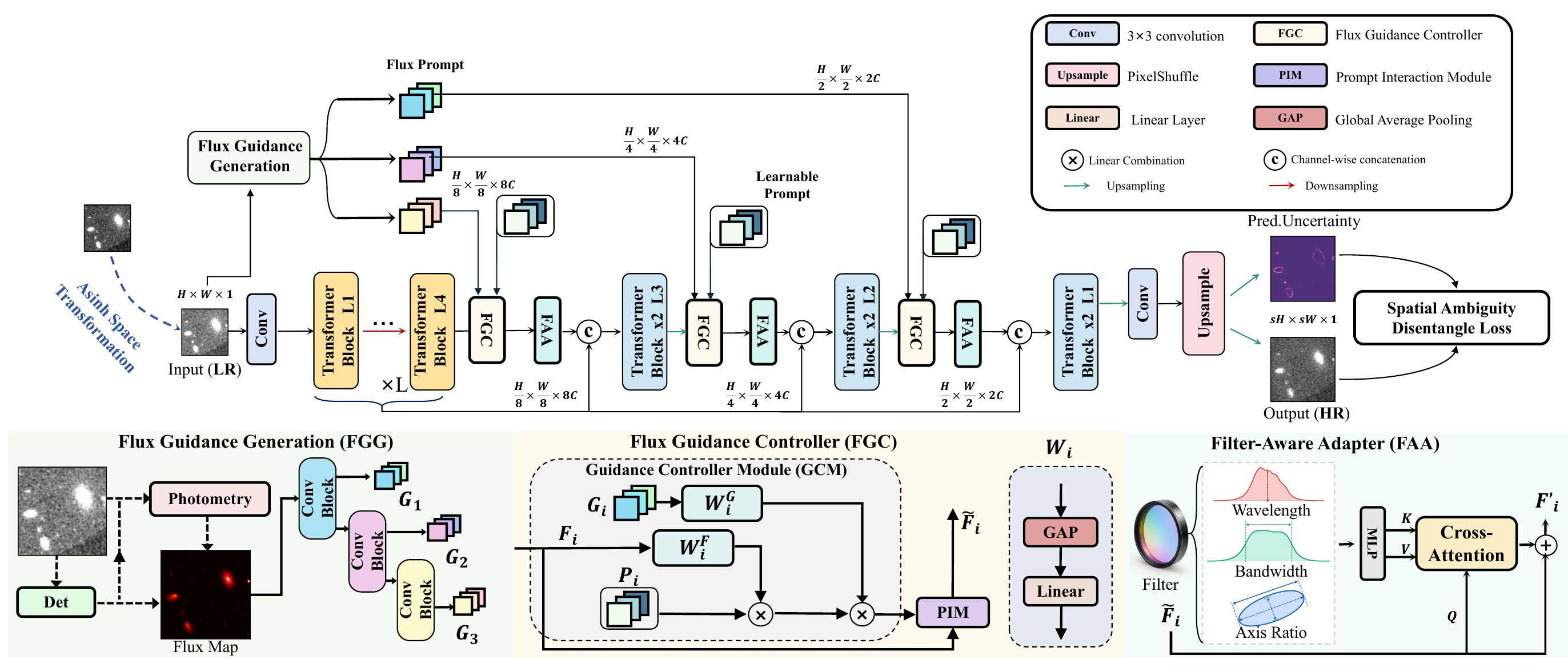}
  \caption{Overview of the proposed FISR++ framework. The LR input is first transformed into the asinh intensity space and processed by an encoder and a Flux Guidance Generation (FGG) module. The extracted flux guidance is incorporated into multi-scale features through Flux Guidance Controllers (FGCs), while the Filter-Aware Adapter (FAA) integrates physical observation priors to enhance feature representations. The enhanced features are progressively decoded and upsampled to reconstruct the HR image. During training, a pixel-wise uncertainty map is also predicted for the  Spatial Ambiguity Disentangle Loss~(SADL).}
    \label{fig3method}
\end{figure*}

%% file: sec/4_Flux-guided_model_optim.tex
\section{Method}
\label{sec:Flux-guided model optim}

\subsection{Overview}
The overall pipeline of our Flux-Invariant Super-Resolution Plus~(FISR++) is shown in Fig.~\ref{fig3method}. The low-resolution input image is first mapped to the asinh intensity space and then processed through two parallel paths, followed by a reconstruction stage. 
The first one is the Flux Guidance Generation~(FGG) module of FISR~\cite{star}, which extracts flux information as multi-scale flux guidance representations. The second is an encoder block consisting of a convolution operation and multiple transformer blocks, which represent the input image as multi-scale feature maps. Each flux guidance is sent into a scale-wise Flux Guidance Controller (FGC) to enhance the encoded feature maps. This scheme highlights regions with significant flux to guide the network’s focus toward astrophysically relevant structures. These features are then processed by a Filter-Aware Adapter~(FAA) to incorporate filter-related physical priors for cross-filter reconstruction. Subsequently, multiple decoder transformer blocks progressively process these features and finally upsample them into the output image. During training, we employ the proposed Spatial Ambiguity Disentangle Loss~(SADL) to improve reconstruction robustness.


\subsection{Asinh Intensity Space}
We adopt an inverse hyperbolic sine (asinh) transformation to represent astronomical images in a compressed intensity space, inspired by its applications in astronomical photometry and image visualization~\cite{lupton2004preparing,stark2018psfgan}. Specifically, we transform both the LR inputs into the asinh intensity space for model training and optimization. For an input image $I'_{\mathrm{lr}}$, the transformation is defined as:
\begin{equation}
I_{\mathrm{lr}} =
\frac{\operatorname{asinh}(I'_{\mathrm{lr}}/\beta)}{\operatorname{asinh}(1/\beta)},
\end{equation}
where $\operatorname{asinh}(x)=\ln\left(x+\sqrt{x^2+1}\right)$ is the inverse hyperbolic sine function, and $\beta$ controls the degree of intensity compression, we set it to 0.1.
As a result, the transformed input $I_{\mathrm{lr}}$ enhances the relative intensities of faint stellar sources, reducing the dominance of bright stars in pixel-wise supervision and increasing the contribution of faint sources during training.

\subsection{Flux Guidance Generation}
\label{sec:fgg}
We propose the (Flux Guidance Generation) FGG module to introduce the flux information. Specifically, given an input image, FGG first represents flux information for every celestial object in the input image into a flux map, then transfers the flux map as multi-scale features as guidance which will then be used in the next FGC module to enhance multi-scale feature maps.

To generate the flux map, FGG also computes flux by detection. It detects celestial objects first and obtains a bounding box for each object. Then the photometry process is operated to derive object fluxes. The entire detection and photometry process is real-time. With these, FGG generates the map following the idea of drawing bounding boxes on a white background and corresponding flux value in each bounding box region. 

In practice, because the 'bounding box' of a celestial object is essentially a rotatable ellipse rather than a rectangle in standard object detection, the drawing scheme FGG has some modifications. Specifically, for each ellipse, FGG puts a rotatable Gaussian Kernel at the center location and adjusts the Gaussian standard deviation based on the ellipse size. Then, FGG multiplies the object flux value directly with the rotated Gaussian kernel to modulate the kernel value. Finally, FGG draws all these kernels together as the final flux map.

Finally, a multi-scale convolutional block transforms the flux map into a corresponding flux guidance representation. In our architecture, we employ three such blocks at different feature levels to generate a hierarchy of flux guidance, denoted as $\{{\mathbf{G}}_1, {\mathbf{G}}_2, {\mathbf{G}}_3\}$. The level 3 is set to align the block number in the decoder, following common setting in a popular SR baseline~\cite{potlapalli2023promptir}. 

\subsection{Flux Guidance Controller}
Based on the flux guidance produced by the FGG module, the (Flux Guidance Controller) FGC interacts such guidance with the encoder features, generating the enhanced features. The FGC is scale-wised and for the $i$-th scale FGC, its interaction pipeline is represented as follows:
\begin{equation}
\hat{\mathbf{F}}_i = \text{PIM}_i\left(\text{GCM}_i\left(\mathbf{P}_i, \mathbf{F}_i,\mathbf{G}_i\right),\mathbf{F}_i\right).
\end{equation}
It shows that the enhanced feature $\hat{\mathbf{F}}_i\in \mathcal{R}^{H \times W \times C}$ is derived from a guidance-controlled feature and an original feature $\mathbf{F}_i\in \mathcal{R}^{H \times W \times C}$ by a prompt interaction module function $\text{PIM}(\cdot,\cdot)$, which is set same as combines image features with guidance-controlled feature and dynamically adjusts the input features through a transformer block. following PromptIR~\cite{potlapalli2023promptir}. The guidance-controlled features are computed by a guidance controller module function $\text{GCM}(\cdot,\cdot,\cdot)$ that takes a learnable prompt $\mathbf{P}_i$, the flux guidance component ${\mathbf{G}}_i\in\mathcal{R}^{H \times W \times C}$ and the encoded feature $\mathbf{F}_i$ as inputs. The learnable prompt $\mathbf{P}_i\in \mathcal{R}^{H \times W \times C\times K}$ contains $K$ learnable patterns expected to represent blind property in the image restoration process~\cite{potlapalli2023promptir}. The detail of GCM is as follows:
\begin{equation}
\text{GCM}_i\left(\mathbf{P}_i, \mathbf{F}_i,\mathbf{G}_i\right) = \sum _{k\in K}W^{F}_i(\mathbf{F}_i)\odot W^{G}_i(\mathbf{G}_i)\odot \mathbf{P}_i,
\end{equation}
where $W_i$ means learnable modules, consisting of a global average pooling and a linear layer, takes input corresponding features and derives a weight with the size of $1\times1\times1\times K$ to indicate the importance of the learnable prompt patterns. $\odot$ means Hadamard product with dimension broadcast while $\sum _{k\in K}$ means sum the last dimension of the multiplied features with the size of $H \times W \times C\times K$ to derive the final output features with the size of $H \times W \times C$.

\subsection{Filter-Aware Adapter}
We propose a Filter-Aware Adapter~(FAA) to incorporate continuous physical priors into image features for cross-filter reconstruction. Specifically, we obtain filter-related physical parameters $\mathbf{m}_f$ from the observation metadata, where $\mathbf{meta}_f$ includes wavelength, bandwidth, and axis ratios. We then normalize these parameters and encode them into filter tokens $\mathbf{P}_f$ through an MLP.

Given the enhanced feature $\hat{\mathbf{F}}_i$ from FGC and the filter tokens $\mathbf{P}_f$, we employ a multi-head cross-attention mechanism to incorporate the physical priors into image features at different scales. The image features serve as queries, while the filter tokens provide keys and values:
\begin{equation}
\mathbf{Z}_i = \mathrm{Softmax}
\left(
\frac{
W_q(\hat{\mathbf{F}}_i)
W_k(\mathbf{P}_f)^T
}
{\sqrt{d}}
\right)
W_v(\mathbf{P}_f),
\end{equation}
where $W_q(\cdot)$, $W_k(\cdot)$, and $W_v(\cdot)$ denote learnable projection layers for query, key, and value, respectively, and $d$ is the key dimension. The attended features are projected back to the original channel dimension and fused with $\hat{\mathbf{F}}_i$ through a residual connection:
\begin{equation}
\mathbf{F}'_i = \hat{\mathbf{F}}_i + W_o(\mathbf{Z}_i),
\end{equation}
where $W_o(\cdot)$ denotes the learnable output projection layer.
Since different filters exhibit distinct spectral characteristics and spatial sampling conditions, incorporating these physical priors enables FISR++ to adapt its image representations to different observations, improving generalization to unseen filters.

\begin{figure}
  \centering
  \label{map}
  \includegraphics[width=1.0\columnwidth]{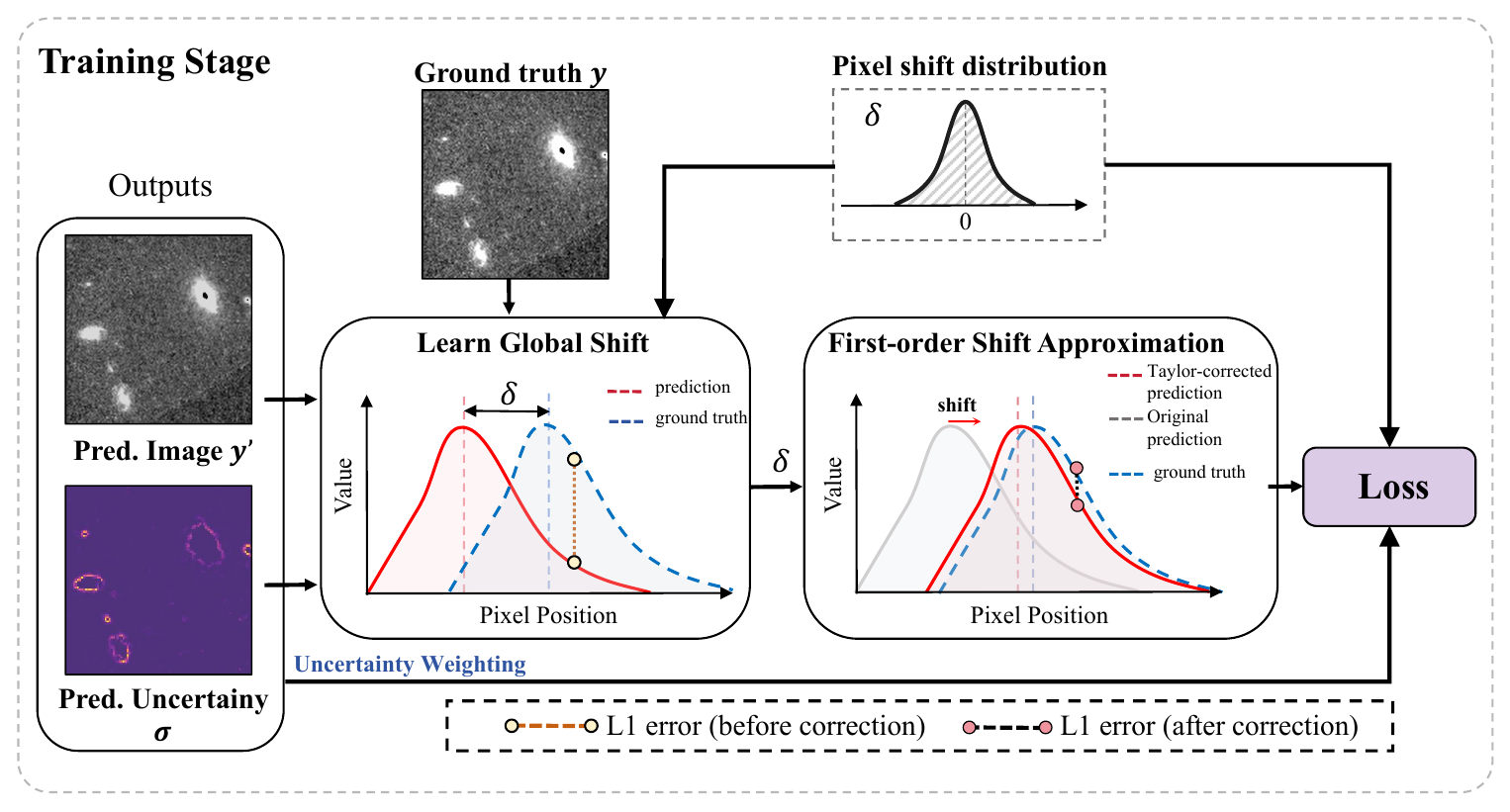}
   \caption{Illustration of the Spatial Ambiguity Disentanglement Loss (SADL). During training, SADL learns a global shift $\boldsymbol{\delta}$ under a Gaussian prior and applies a first-order Taylor correction to the prediction. The curves illustrate pixel-value profiles, where the vertical distances between the prediction and ground truth represent pixel-wise L1 errors before and after shift correction, weighted by the predicted uncertainty.}
    \label{map}
\end{figure}

\subsection{Spatial Ambiguity Disentangle Loss}
To model spatial ambiguity in the reconstruction of high-frequency stellar structures, we propose a Spatial Ambiguity Disentanglement Loss (SADL), as shown in Fig.~\ref{map}. Given the reconstructed image $y'\in\mathbb{R}^{H\times W}$, the HR ground truth $y\in\mathbb{R}^{H\times W}$, and the predicted pixel-wise uncertainty $\sigma\in\mathbb{R}^{H\times W}$, SADL is formulated as:
\begin{equation}
\label{eq:map_loss}
\small
\begin{aligned}
\mathcal{L}_{\mathrm{SADL}}
=\frac{1}{HW}\bigg[
&
\begin{aligned}
\sum_{v=0}^{H-1}\sum_{u=0}^{W-1}
\Bigg(
&\underbrace{\frac{\sqrt{2}}{\sigma(u,v)}
\Big|y(u,v)-y'(u+\Delta u,v+\Delta v)\Big|}_{\text{Uncertainty-weighted pixel error}}
\Bigg)
\end{aligned}
\\[6pt]
&+\underbrace{\log\sigma(u,v)}_{\substack{\text{Uncertainty} \\ \text{Regularization}}}+\underbrace{
\frac{\|\boldsymbol{\delta}\|_2^2}
{2\widetilde{\sigma}^{2}}
}_{\substack{\text{Spatial Shift} \\ \text{Regularization}}}
\bigg],\ \ \ \text{where}\ \delta=(\Delta u, \Delta v).
\end{aligned}
\end{equation}
$(u,v)$ denotes the pixel coordinates and $\delta=(\Delta u, \Delta v)$ is a learnable global shift pair. The coefficient $\sqrt{2}/\sigma$ before the L1 error term serves as an uncertainty-dependent weight. Specifically, when the predicted uncertainty $\sigma(u,v)$ is large, the coefficient $\sqrt{2}/\sigma(u,v)$ becomes smaller, reducing the contribution of the corresponding reconstruction error. Meanwhile, the uncertainty regularization term penalizes excessively large uncertainty estimates, preventing the model from trivially reducing the reconstruction error by unboundedly increasing the predicted uncertainty. The spatial shift regularization term plays a similar role to $\delta$, whose regularization strength is controlled by a hyperparameter $\widetilde{\sigma}$. Based on this, to make $y'(u+\Delta u, v+\Delta v)$ differentiable with respect to both $y'$ and $\boldsymbol{\delta}$, we approximate it via its first-order Taylor expansion:
\begin{equation}
\label{eq:shift_taylor}
y'(u+\Delta u,v+\Delta v)
\approx
y'(u,v)+\nabla y'(u,v)^\top\boldsymbol{\delta}.
\end{equation}
The first-order term $\nabla y'(u,v)^\top\boldsymbol{\delta}$ approximates the intensity variation induced by the global shift. 

The entire SADL is theoretically derived from a maximum a posteriori (MAP) idea to optimize pixel shift, which could be formulated as:
\begin{equation}
\label{eq:map_estimation}
\begin{aligned}
p(\boldsymbol{\delta}\mid y,y',\sigma)
&=
\frac{
p(y\mid y'_{\boldsymbol{\delta}},\sigma)
p(\boldsymbol{\delta})
}{
p(y\mid y',\sigma)
}\\
&\propto
p(y\mid y'_{\boldsymbol{\delta}},\sigma)
p(\boldsymbol{\delta}),
\end{aligned}
\end{equation}
The denominator $p(y\mid y'_{\boldsymbol{\delta}},\sigma)$ can be omitted during optimization, due to the approximately uniform or slowly varying sky backgrounds commonly observed in local regions of astronomical images~\cite{SExtractor,ProFound}. So, the maximization could be re-written under log space as :
\begin{equation}
\begin{aligned}
\text{max}\ \log p(\boldsymbol{\delta}\mid y,y',\sigma)
&=\text{max}
\log p(y\mid y'_{\boldsymbol{\delta}},\sigma)
+\log p(\boldsymbol{\delta}),
\end{aligned}
\end{equation}
where $p(y\mid y'_{\boldsymbol{\delta}},\sigma)$ is modeled by our reconstruction network and $p(\boldsymbol{\delta})$ is assumed as a zero-mean Gaussian prior, $\boldsymbol{\delta}\sim\mathcal{N}(\mathbf{0},\widetilde{\sigma}^2\mathbf{I}_2)$. It transfers to the spatial shift regularization term $\|\boldsymbol{\delta}\|_2^2/(2\widetilde{\sigma}^2)$ constrains the shift magnitude and discourages extreme displacements.

By jointly accounting for pixel-wise uncertainty and latent sub-pixel shifts, SADL reduces sensitivity to positional deviations while preserving the reconstruction constraints on stellar structures. The shift is estimated only during training and is not applied to the reconstructed image at inference. Further implementation details are provided in the supplementary material.

\subsection{Flux Consistency Loss}
The idea of the Flux Consistency Loss is to train the model to generate flux consistent prediction. 
We propose a simple yet effective scheme to achieve this goal by highlighting flux-related regions. We first operate the flux map generation scheme described in Section~\ref{sec:fgg} on the ground-truth image, denoted as $M$. Additionally, the HR ground-truth image is also transformed into the asinh intensity space to match the reconstructed prediction. They we use this map to weighted the pixel wised supervision as follows:

\begin{equation}
\mathcal{L}_{\text{flux}}(I_{\text{pred}}, I_{\text{gt}}) = \sum_{x,y} M(x,y)\cdot\left|I_{\text{pred}}(x,y) - I_{\text{gt}}(x,y) \right|.
\end{equation}

Combined with the proposed Spatial Ambiguity Disentangle Loss, the overall training objective is defined as:
\begin{equation}
\mathcal{L}_{\text{total}}
=
\mathcal{L}_{\text{SADL}}
+
\lambda_{\text{flux}}\mathcal{L}_{\text{flux}},
\end{equation}
where \(\lambda\) balances terms. This loss takes into account both traditional regression supervision and flux consistency constraints, which brings more physically reliable ASR results.

\begin{table*}[!t]
\centering
\caption{ASR performance on seven unseen STAR++ filters at $\times 2$.
The interpolation subset comprises F435W, F555W, F606W, and F625W,
while the extrapolation subset comprises F218W, F225W, and F850LP.
STAR and STAR++ indicate the training datasets.}
\label{tab:main_results_x2}

\footnotesize
\setlength{\tabcolsep}{7pt}
\renewcommand{\arraystretch}{1.22}

\begin{tabular*}{\textwidth}{
@{\extracolsep{\fill}}llccccc@{}
}
\toprule

\multirow{2}{*}{\textbf{Training set}}
& \multirow{2}{*}{\textbf{Method}}
& \multicolumn{3}{c}{\textbf{PSNR (dB)} $\uparrow$}
& \multirow{2}{*}{\textbf{SSIM} $\uparrow$}
& \multirow{2}{*}{\textbf{FE} $\downarrow$} \\

\cmidrule(lr){3-5}

& & \textbf{Interp.}
& \textbf{Extrap.}
& \textbf{Overall} & & \\

\midrule

-- & Bicubic
& 29.72 & 23.53 & 27.06 & 0.7585 & 2.8908 \\

\midrule

\multirow{6}{*}{STAR}
& EDSR
& 31.94 & 24.16 & 28.61 & 0.7934 & 1.2957 \\

& RCAN
& 33.86 & 26.79 & 30.83 & 0.8201 & 0.7404 \\

& SwinIR
& 34.79 & 26.95 & 31.43
& \cellcolor{FISRlight}0.8264
& 0.7141 \\

& HAT
& 35.14 & 26.76 & 31.55 & 0.8253 & 0.6205 \\

& FISR
& \cellcolor{FISRlight}35.57
& \cellcolor{FISRlight}27.34
& \cellcolor{FISRlight}32.04
& 0.8216
& \cellcolor{FISRlight}0.6027 \\

\rowcolor{FISRppLight}
\cellcolor{white}
& FISR++
& \textbf{36.89}
& \textbf{28.63}
& \textbf{33.35}
& \textbf{0.8280}
& \textbf{0.5888} \\

\midrule

\multirow{6}{*}{STAR++}
& EDSR
& 36.08 & 29.39 & 33.21 & 0.8391 & 0.7846 \\

& RCAN
& 36.74 & 30.73 & 34.16 & 0.8414 & 0.7247 \\

& SwinIR
& 37.12 & 30.35 & 34.22 & 0.8522 & 0.6607 \\

& HAT
& \cellcolor{FISRlight}37.93
& \cellcolor{FISRlight}31.28
& \cellcolor{FISRlight}35.08
& \cellcolor{FISRlight}0.8585
& 0.5617 \\

& FISR
& 37.22 & 30.46 & 34.32 & 0.8542
& \cellcolor{FISRlight}0.5304 \\

\rowcolor{FISRppLight}
\cellcolor{white}
& FISR++
& \textbf{38.05}
& \textbf{31.91}
& \textbf{35.42}
& \textbf{0.8623}
& \textbf{0.4341} \\

\bottomrule
\end{tabular*}
\end{table*}

\begin{table*}[!t]
\centering
\caption{ASR performance on unseen STAR++ filters at $\times 4$.
STAR and STAR++ indicate the training datasets.}
\label{tab:main_results_x4}

\footnotesize
\setlength{\tabcolsep}{7pt}
\renewcommand{\arraystretch}{1.22}

\begin{tabular*}{\textwidth}{
@{\extracolsep{\fill}}llccccc@{}
}
\toprule

\multirow{2}{*}{\textbf{Training set}}
& \multirow{2}{*}{\textbf{Method}}
& \multicolumn{3}{c}{\textbf{PSNR (dB)} $\uparrow$}
& \multirow{2}{*}{\textbf{SSIM} $\uparrow$}
& \multirow{2}{*}{\textbf{FE} $\downarrow$} \\

\cmidrule(lr){3-5}

& & \textbf{Interp.}
& \textbf{Extrap.}
& \textbf{Overall} & & \\

\midrule

-- & Bicubic
& 26.23 & 20.41 & 23.74 & 0.6523 & 4.4267 \\

\midrule

\multirow{6}{*}{STAR}
& EDSR
& 30.84 & 23.44 & 27.67 & 0.7165 & 1.0674 \\

& RCAN
& 31.53 & 23.67 & 28.16 & 0.7234 & 0.8701 \\

& SwinIR
& 31.68 & 24.03 & 28.40 & 0.7283 & 0.8273 \\

& HAT
& 31.99 & 24.03 & 28.58 & 0.7328
& \cellcolor{FISRlight}0.7659 \\

& FISR
& \cellcolor{FISRlight}32.24
& \cellcolor{FISRlight}24.11
& \cellcolor{FISRlight}28.76
& \cellcolor{FISRlight}0.7345
& 0.7670 \\

\rowcolor{FISRppLight}
\cellcolor{white}
& FISR++
& \textbf{32.53}
& \textbf{25.75}
& \textbf{29.62}
& \textbf{0.7653}
& \textbf{0.7540} \\

\midrule

\multirow{6}{*}{STAR++}
& EDSR
& 33.08 & 26.92 & 30.44 & 0.7573 & 0.8683 \\

& RCAN
& 33.69 & 26.83 & 30.75 & 0.7634 & 0.7639 \\

& SwinIR
& 34.03 & 27.31 & 31.15 & 0.7681 & 0.7559 \\

& HAT
& \cellcolor{FISRlight}34.44
& 27.76
& \cellcolor{FISRlight}31.58
& \cellcolor{FISRlight}0.7773
& 0.7494 \\

& FISR
& 34.09
& \cellcolor{FISRlight}28.10
& 31.52
& 0.7736
& \cellcolor{FISRlight}0.7257 \\

\rowcolor{FISRppLight}
\cellcolor{white}
& FISR++
& \textbf{35.35}
& \textbf{29.01}
& \textbf{32.63}
& \textbf{0.7805}
& \textbf{0.6803} \\

\bottomrule
\end{tabular*}
\end{table*}

%% file: sec/5_experiment.tex
\section{Experiment}
\label{sec:experiment}










\subsection{Experimental Setup}
\noindent \textbf{Dataset}. 
We consider two training settings, models trained on the original STAR dataset and models trained on the proposed STAR++ dataset. Both are evaluated on the unseen-filter test sets of STAR++ to assess cross-filter generalization and the benefits of multi-filter training. 


\noindent \textbf{Implementation Details}. 
All models are trained for 24 epochs with a batch size of 16 on 8 Ascend 910B2 NPUs. The initial learning rate is set to $2\times10^{-4}$ and decayed by a factor of 0.01 at the 12th epoch. A linear warm-up strategy is applied during the first epoch to stabilize training. For image degradation, we set the Gaussian PSF parameter $\sigma \in [0.8, 1.2]$ and the Airy PSF radius $r \in [1.9, 2.2]$ pixels based on the FWHM~\cite{sirianni2005photometric}, which measures the blur width at half its peak intensity, to approximate the actual HST WFC/ACS F814W filter observations where the blur is characterized by its FWHM. The standard deviation of the Gaussian prior on the pixel shift is set to $\widetilde{\sigma}=0.15$.

\begin{table*}[t]
\centering
\caption{Ablation studies of different components under the
$\times 2$ super-resolution setting.
UL and SSR denote Uncertainty Loss and Spatial Shift
Regularization, respectively.}
\label{tab:ablation_all_x2}

\footnotesize
\setlength{\tabcolsep}{5pt}
\renewcommand{\arraystretch}{1.15}

\resizebox{0.90\textwidth}{!}{
\begin{tabular}{cc|ccc|cccc|ccc}
\toprule

\multirow{3}{*}{\textbf{Dataset}}
& \multirow{3}{*}{\textbf{ID}}
& \multicolumn{3}{c|}{\textbf{FISR}}
& \multicolumn{4}{c|}{\textbf{FISR++}}
& \multicolumn{3}{c}{\textbf{Performance}} \\

\cmidrule(lr){3-5}
\cmidrule(lr){6-9}
\cmidrule(lr){10-12}

&
& \multirow{2}{*}{\textbf{FGG}}
& \multirow{2}{*}{\textbf{FGC}}
& \multirow{2}{*}{\textbf{FCL}}
& \multirow{2}{*}{\textbf{FAA}}
& \multicolumn{2}{c}{\textbf{SADL}}
& \multirow{2}{*}{\textbf{Asinh}}
& \multirow{2}{*}{\textbf{PSNR$\uparrow$}}
& \multirow{2}{*}{\textbf{SSIM$\uparrow$}}
& \multirow{2}{*}{\textbf{FE$\downarrow$}} \\

\cmidrule(lr){7-8}

&
& & & &
& \textbf{UL}
& \textbf{SSR}
& & & & \\

\midrule

\multirow{6}{*}{STAR}
& (a) & - & - & - & - & - & - & -
& 37.77 & 0.8288 & 0.7022 \\

& (b) & - & - & \checkmark & - & - & - & -
& 37.86 & 0.8283 & 0.6389 \\

& (c) & \checkmark & - & - & - & - & - & -
& 37.71 & 0.8286 & 0.7572 \\

& (d) & \checkmark & - & \checkmark & - & - & - & -
& 37.87 & 0.8301 & 0.7467 \\

& (e) & \checkmark & \checkmark & - & - & - & - & -
& 37.85 & 0.8302 & 0.6527 \\

& (f) & \checkmark & \checkmark & \checkmark
& - & - & - & -
& \textbf{37.88} & \textbf{0.8311} & \textbf{0.5739} \\

\midrule

\multirow{6}{*}{STAR++}
& (g) & \checkmark & \checkmark & \checkmark
& - & - & - & -
& 34.32 & 0.8542 & 0.5304 \\

& (h) & \checkmark & \checkmark & \checkmark
& \checkmark & - & - & -
& 34.64 & 0.8571 & 0.5207 \\

& (i) & \checkmark & \checkmark & \checkmark
& - & \checkmark & - & -
& 34.59 & 0.8574 & 0.4714 \\

& (j) & \checkmark & \checkmark & \checkmark
& - & \checkmark & \checkmark & -
& 34.87 & 0.8603 & 0.4549 \\

& (k) & \checkmark & \checkmark & \checkmark
& \checkmark & \checkmark & \checkmark & -
& 35.27 & 0.8614 & 0.4536 \\

& (l) & \checkmark & \checkmark & \checkmark
& \checkmark & \checkmark & \checkmark & \checkmark
& \textbf{35.42} & \textbf{0.8623} & \textbf{0.4341} \\

\bottomrule
\end{tabular}
}
\end{table*}

\begin{table}[t]
\centering
\caption{Sensitivity analysis of the Gaussian prior standard deviation $\widetilde{\sigma}$ in SADL. The best results are highlighted in bold.}
\label{tab:prior_sensitivity}
\footnotesize
\setlength{\tabcolsep}{12pt}
\renewcommand{\arraystretch}{1.15}
\begin{tabular}{cccc}
\toprule
$\widetilde{\sigma}$
& PSNR (dB)$\uparrow$
& SSIM$\uparrow$
& FE$\downarrow$ \\
\midrule
0.05 & 35.37 & 0.8618 & 0.4378 \\
0.10 & 35.39 & 0.8619 & 0.4347 \\
\textbf{0.15} & \textbf{35.42} & \textbf{0.8623} & \textbf{0.4341} \\
0.20 & 35.41 & 0.8622 & 0.4343 \\
0.50 & 34.94 & 0.8607 & 0.4401 \\
\bottomrule
\end{tabular}
\end{table}

\begin{table*}[t]
\centering
\caption{Comparison of different methods with and without FCL under $\times 2$ and $\times 4$ ASR settings on STAR and STAR++. Bold entries indicate that the setting with FCL (w/) improves over the setting without FCL (w/o) for the same metric and method.}
\label{tab:flux_loss_cleaned}
\footnotesize
\setlength{\tabcolsep}{5pt}
\renewcommand{\arraystretch}{1.12}

\resizebox{\textwidth}{!}{
\begin{tabular}{c|c|cccccc|c}
\toprule

\multirow{2}{*}{\textbf{Scale}}
& \multirow{2}{*}{\textbf{Flux Loss}}
& \multicolumn{6}{c|}{\textbf{STAR}}
& \multicolumn{1}{c}{\textbf{STAR++}} \\

\cmidrule(lr){3-8}
\cmidrule(lr){9-9}

&
& \textbf{EDSR~\cite{lim2017enhanced}}
& \textbf{RCAN~\cite{zhang2018image}}
& \textbf{SwinIR~\cite{liang2021swinir}}
& \textbf{HAT~\cite{chen2023activating}}
& \textbf{FISR}
& \textbf{FISR++}
& \textbf{FISR++} \\

\midrule

\multirow{6}{*}{$\times 2$}

& PSNR$\uparrow$ (\textcolor{red}{w/o})
& 35.74 & 37.46 & \textbf{37.60} & 37.91 & 38.01 & 38.09
& 35.28 \\

& SSIM$\uparrow$ (\textcolor{red}{w/o})
& 0.8086 & 0.8277 & 0.8279 & 0.8296 & 0.8313 & 0.8346
& 0.8605 \\

& FE$\downarrow$ (\textcolor{red}{w/o})
& 1.3249 & 0.7451 & 0.7593 & 0.6918 & 0.5775 & 0.5469
& 0.4410 \\

\cmidrule(lr){2-9}

& PSNR$\uparrow$ (\textcolor{green}{w/})
& \textbf{35.89} & \textbf{37.89} & 37.13
& \textbf{37.94} & \textbf{38.27} & \textbf{38.83}
& \textbf{35.42} \\

& SSIM$\uparrow$ (\textcolor{green}{w/})
& \textbf{0.8092} & \textbf{0.8286} & \textbf{0.8281}
& \textbf{0.8303} & \textbf{0.8334} & \textbf{0.8384}
& \textbf{0.8623} \\

& FE$\downarrow$ (\textcolor{green}{w/})
& \textbf{1.2420} & \textbf{0.5914} & \textbf{0.6767}
& \textbf{0.5738} & \textbf{0.5585} & \textbf{0.5012}
& \textbf{0.4341} \\

\midrule

\multirow{6}{*}{$\times 4$}

& PSNR$\uparrow$ (\textcolor{red}{w/o})
& 33.21 & \textbf{35.70} & 35.11 & 35.87 & 36.23 & 36.91
& 32.47 \\

& SSIM$\uparrow$ (\textcolor{red}{w/o})
& 0.7201 & 0.7263 & 0.7258 & 0.7276 & 0.7233 & 0.7628
& 0.7765 \\

& FE$\downarrow$ (\textcolor{red}{w/o})
& 1.681 & 1.2355 & 1.3705 & 1.1163 & 1.0861 & 0.9948
& 0.6811 \\

\cmidrule(lr){2-9}

& PSNR$\uparrow$ (\textcolor{green}{w/})
& \textbf{33.45} & 35.43 & \textbf{35.32} & \textbf{36.16} & \textbf{36.46} & \textbf{37.17} & \textbf{32.63} \\

& SSIM$\uparrow$ (\textcolor{green}{w/})
& \textbf{0.7295} & \textbf{0.7310} & \textbf{0.7338} & \textbf{0.7364} & \textbf{0.7352} & \textbf{0.7662} & \textbf{0.7805} \\

& FE$\downarrow$ (\textcolor{green}{w/})
& \textbf{1.2459} & \textbf{1.0055} & \textbf{0.9234} & \textbf{0.9299} & \textbf{0.9625} & \textbf{0.6803} 
& \textbf{0.6803} \\

\bottomrule
\end{tabular}
}
\end{table*}

\noindent \textbf{Evaluation Protocols}. 
Following prior work, we adopt Peak Signal-to-Noise Ratio (PSNR)~\cite{antkowiak2000final} and Structural Similarity Index (SSIM)~\cite{wang2004image} as standard metrics for evaluating reconstruction quality. In addition, we use the proposed FE to evaluate the physical fidelity of the reconstructed images by measuring their consistency with astrophysical properties. We do not adopt commonly used perceptual metrics for image super-resolution, such as LPIPS~\cite{zhang2018unreasonable}, since these metrics rely on deep models pretrained on natural-image datasets such as ImageNet~\cite{deng2009imagenet,simonyan2014very,he2016deep}. Due to the substantial domain gap between natural and astronomical imagery, such pretrained representations may not provide an appropriate measure of perceptual similarity for astronomical images.

\subsection{Quantitative and Qualitative Results}
We compare FISR++ with Bicubic interpolation and representative state-of-the-art SR methods, including CNN-based EDSR~\cite{lim2017enhanced} and RCAN, Transformer-based SwinIR~\cite{liang2021swinir} and HAT~\cite{chen2023activating}, and the astronomical SR method FISR~\cite{star} are also popular in the ASR topic~\cite{miao2024astrosr,reddy2024difflense,FluxFlow}, which is also an important factor that we choose them to compare. As shown in Tab.~\ref{tab:main_results_x2} and Tab.~\ref{tab:main_results_x4}, we compare models trained on STAR and STAR++ using the same unseen-filter test sets of STAR++, and the results show that our approach outperforms existing methods across different evaluation criteria.

Compared with their STAR-trained counterparts, models trained on STAR++ generally achieve higher overall PSNR on unseen filters. Specifically, EDSR improves from 28.61 to 33.21 dB at $\times 2$ and from 27.67 to 30.44 dB at $\times 4$. HAT achieves gains of 3.53 dB and 3.00 dB at the two scales, respectively, while FISR improves by 2.28 dB and 2.76 dB. 

We further analyze generalization to unseen interpolation and extrapolation filters. At $\times 2$, FISR++ achieves PSNR values of 38.05 dB and 31.91 dB on the interpolation and extrapolation subsets, respectively, outperforming FISR by 0.83 dB and 1.45 dB. At $\times 4$, the corresponding values reach 35.35 dB and 29.01 dB, with improvements of 1.26 dB and 0.91 dB over FISR. Overall, these results indicate that training on STAR++ generally improves model generalization to unseen filters, while FISR++ consistently outperforms FISR across both interpolation and extrapolation settings, demonstrating its superior cross-filter generalization capability.

Among models trained on STAR++, FISR++ achieves the highest overall PSNR of 35.42 dB and 32.63 dB at $\times 2$ and $\times 4$, respectively. Compared with FISR under the same training setting, FISR++ improves PSNR by 1.10 dB and 1.11 dB, while reducing FE from 0.5304 to 0.4341 and from 0.7257 to 0.6803 at the two scales, respectively. These results demonstrate the effectiveness of FISR++ in improving reconstruction accuracy and flux consistency across unseen filters.

\begin{figure*}
  \centering
  \label{fig:4}
  \includegraphics[width=2\columnwidth]{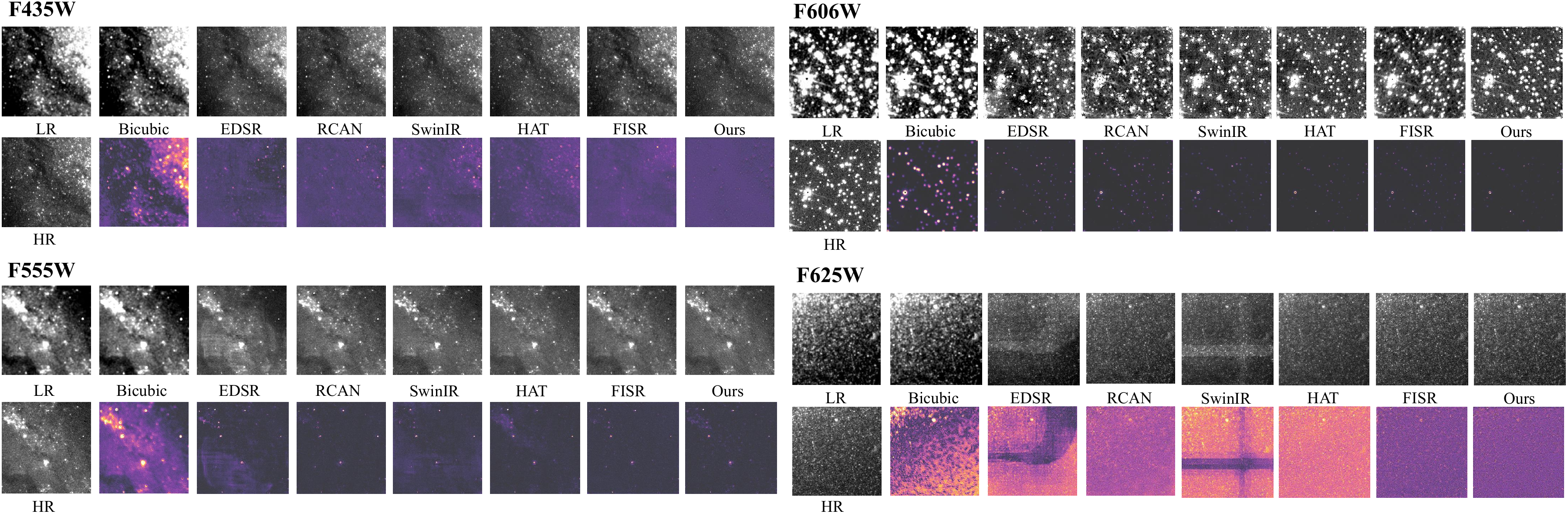}
  \caption{Visualization of SR reconstruction results on unseen interpolation filters. For each filter, the upper row shows the reconstructions from different methods, while the lower row presents the corresponding residual maps.}
    \label{fig:4}
\end{figure*}

\begin{figure*}
  \centering
  \label{fig:5}
  \includegraphics[width=2.0\columnwidth]{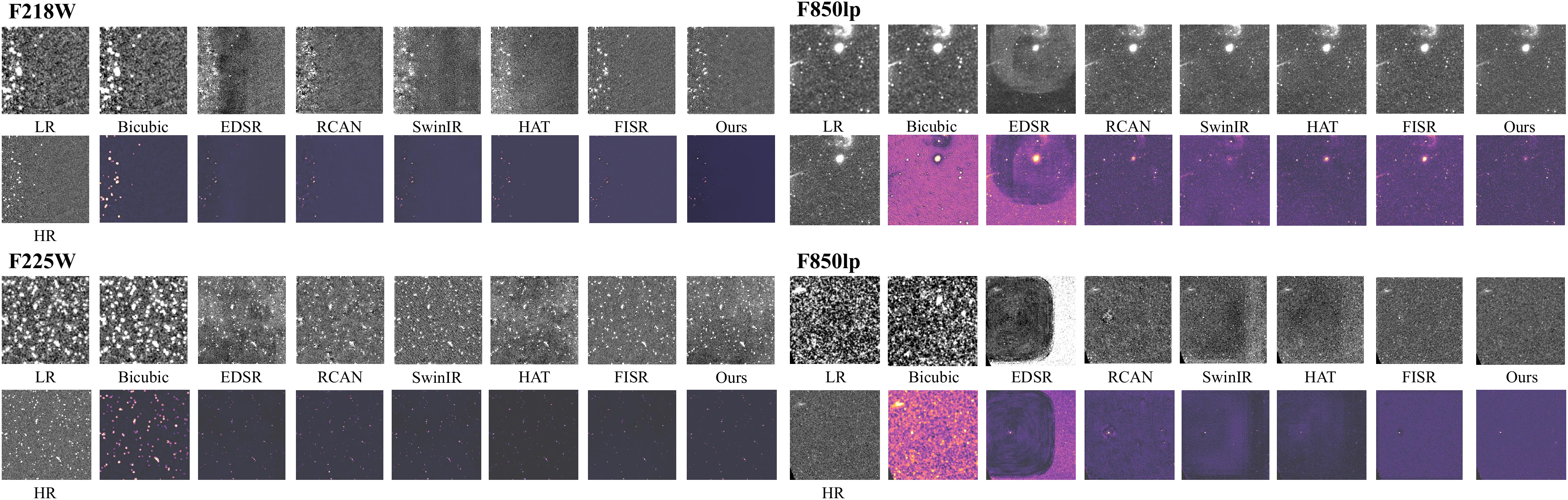}
  \caption{Visualization of SR reconstruction results on unseen extrapolation filters. For each filter, the upper row shows the reconstructions from different methods, while the lower row presents the corresponding residual maps.}
    \label{fig:5}
\end{figure*}

\subsection{Ablation Study}
To evaluate the contribution of each proposed component in our FISR++ framework, we conduct comprehensive ablation studies on the STAR++ dataset under the $\times2$ super-resolution settings.
Tab.~\ref{tab:ablation_all_x2} reports the ablation results of the three conference-version components (FGG, FGC, and FCL) on STAR, followed by the three newly introduced components (FAA, SADL, and Asinh) on STAR++. Specifically, SADL is further decomposed into uncertainty loss (UL) and spatial shift regularization (SSR) to evaluate their individual contributions. We also investigate the sensitivity of SADL to the Gaussian prior standard deviation $\widetilde{\sigma}$, as reported in Tab.~\ref{tab:prior_sensitivity}.

\textbf{Effect of FGG, FGC, and FCL: }We briefly revisit the three components introduced in the conference version. As shown in configurations (a)–(f), although individual combinations do not consistently improve all metrics, integrating all three components achieves the best overall performance on STAR, improving PSNR from 37.77 to 37.88 dB and reducing FE from 0.7022 to 0.5739. These results confirm the combined effectiveness of the original flux-guidance design.

We further evaluate the effectiveness of FCL across seven model configurations at $\times 2$ and $\times 4$ scales. As shown in Tab.~\ref{tab:flux_loss_cleaned}, incorporating FCL consistently reduces FE across all seven configurations at both scales. At $\times 2$, FE decreases from 0.5775 to 0.5585 for FISR trained on STAR and from 0.4410 to 0.4341 for FISR++ trained on STAR++. At $\times 4$, the corresponding reductions are from 1.0861 to 0.9625 for FISR and from 0.6811 to 0.6803 for FISR++. These results demonstrate that FCL consistently improves photometric fidelity in terms of FE, although the magnitude of improvement varies across models and upsampling scales.

\textbf{Effect of FAA:} Comparing configurations (g) and (h), introducing FAA improves PSNR by 0.32 dB and SSIM by 0.0029, while reducing FE from 0.5304 to 0.5207. These results demonstrate the effectiveness of incorporating filter-specific physical priors into the reconstruction framework.

\textbf{Effect of SADL:} Compared with configuration (g), introducing UL in (i) improves PSNR by 0.27 dB and SSIM by 0.0032, while reducing FE from 0.5304 to 0.4714. Further incorporating spatial shift modeling with regularization in (j) improves PSNR by an additional 0.28 dB and SSIM by 0.0029, while reducing FE to 0.4549. Moreover, jointly incorporating FAA and the complete SADL in (k) improves PSNR by 0.95 dB and SSIM by 0.0072, while reducing FE to 0.4536 compared with (g). These results demonstrate the complementary benefits of pixel-wise uncertainty modeling and spatial shift regularization, with further improvements from filter-aware adaptation.

\textbf{Effect of Asinh:} Comparing configurations (k) and (l), introducing the asinh intensity representation improves PSNR by 0.15 dB and SSIM by 0.0009, while reducing FE from 0.4536 to 0.4341. The complete FISR++ achieves the best overall performance, outperforming the FISR baseline (g) by 1.10 dB in PSNR. These results suggest that compressing the intensity dynamic range helps balance the contributions of bright and faint sources during reconstruction, improving both reconstruction quality and photometric fidelity.

\textbf{Sensitivity Analysis of the Shift Prior:} We investigate the sensitivity of SADL to the prior standard deviation $\widetilde{\sigma}$ by varying it from 0.05 to 0.50. As shown in Table~\ref{tab:prior_sensitivity}, the reconstruction performance remains relatively stable when $\widetilde{\sigma}$ ranges from 0.10 to 0.20. The best performance is achieved at $\widetilde{\sigma}=0.15$, with an PSNR of 35.42 dB, an SSIM of 0.8623, and an FE of 0.4341. When $\widetilde{\sigma}$ increases to 0.50, PSNR decreases by 0.48 dB, indicating that excessively weak shift regularization may degrade reconstruction performance. These results support the choice of $\widetilde{\sigma}=0.15$ as the default setting.

\begin{figure*}
  \centering
  \label{fig:6}
  \includegraphics[width=2.0\columnwidth]{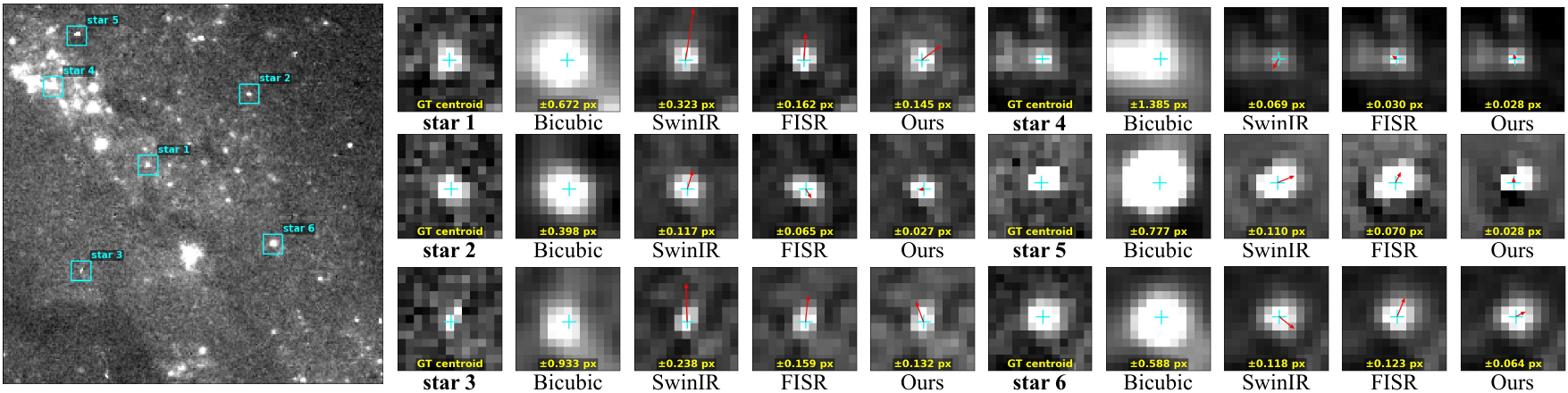}
  \caption{Visualization of stellar centroid shifts under different SR methods. Left: six selected sources marked by boxes. Right: comparison of their centroid offsets across different SR methods.}
    \label{fig:6}
\end{figure*}

\subsection{Quality Analysis}
Fig.~\ref{fig:4} and Fig.~\ref{fig:5} present qualitative comparisons of different SR methods on unseen interpolation and extrapolation filters, respectively. Although existing methods can recover the overall structures of star fields, they often produce noticeable reconstruction residuals, particularly around bright stars and densely populated regions. In contrast, FISR++ produces cleaner residual maps across both settings while preserving stellar structures, demonstrating its effectiveness in reconstructing unseen-filter observations.

To further evaluate pixel shift in reconstructed star fields, Fig.~\ref{fig:6} presents the stellar centroid offsets obtained by different methods. The cyan crosses indicate the ground-truth centroids, while the red lines illustrate the positional deviations. FISR++ achieves smaller centroid offsets than the compared methods for all six selected sources. In particular, its offsets are consistently smaller than those of FISR, suggesting that the proposed SADL helps reduce spatial misalignment during reconstruction. These results provide further evidence that FISR++ not only improves pixel-level reconstruction quality but also better preserves the positional accuracy of individual stellar sources.

%% file: sec/6_conclusion.tex
\section{Conclusion}
\label{sec/6_conclusion}

In this work, we present STAR++, a large-scale, multi-filter benchmark for astronomical super-resolution~(ASR). To address the limited filter diversity and lack of systematic unseen-filter evaluation in existing benchmarks, STAR++ provides approximately 1.7 million flux-consistent LR--HR image pairs across 36 HST filters, with sky-region-disjoint splits supporting the evaluation of cross-filter generalization under both interpolation and extrapolation settings. Building on STAR++, we further propose FISR++, which extends the original flux-guided reconstruction framework with a Filter-Aware Adapter (FAA), an asinh intensity representation, and a Spatial Ambiguity Disentangle Loss~(SADL). Specifically, these components incorporate filter-specific physical information into feature representations, balance the contributions of bright and faint sources during training, and explicitly account for pixel-wise reconstruction uncertainty and sub-pixel spatial misalignment, respectively. Extensive experiments demonstrate that training on STAR++ consistently improves unseen-filter reconstruction across multiple SR architectures. Furthermore, FISR++ achieves the best overall performance among the compared methods, attaining the highest PSNR and SSIM and the lowest Flux Error (FE) at both $\times 2$ and $\times 4$ super-resolution scales.